\documentclass{article}

 \usepackage[preprint]{neurips_2020}

\usepackage[utf8]{inputenc} 
\usepackage[T1]{fontenc}    
\usepackage{hyperref}       
\usepackage{url}            
\usepackage{booktabs}       
\usepackage{amsfonts}       
\usepackage{nicefrac}       
\usepackage{microtype}      
\usepackage{graphicx}
\usepackage{amssymb}
\usepackage{mathtools}
\usepackage{authblk}

\usepackage{booktabs}       
\usepackage{amsfonts}       
\usepackage{nicefrac}       
\usepackage{microtype}      
\usepackage[noend]{algpseudocode}
\usepackage{algorithm,algorithmicx}
\usepackage{hyperref}       
\usepackage{url}            
\usepackage[noend]{algpseudocode}
\usepackage{algorithm,algorithmicx}

\usepackage{booktabs}
\usepackage{array, makecell} 

\newcommand*\sub[1]{_\textrm{\scriptsize #1}}
\newcommand*\Let[2]{\State #1 $\gets$ #2}
\newcommand*\Lets[4]{\State #1 $\gets$ #2, #3 $\gets$ #4}

\newtheorem{definition}{Definition}
\newtheorem{theorem}{Lemma}

\newcommand \nt \textrm 
\title{Radial Autoencoders for Enhanced Anomaly Detection}

\author[1  \thanks{Main contributor of the paper. Work done when Mihai-Cezar was a research data scientist at HES-SO}]{Mihai-Cezar Augustin }
\author[2]{Vivien Bonvin}
\author[1]{Régis Houssou}
\author[1]{Efstratios Rappos}
\author[1  \thanks{  Corresponding author: stephan.robert@hes-so.ch  }]{\\Stephan Robert-Nicoud}

\affil[1]{Haute Ecole de la Suisse Occidentale (HES-SO)}
\affil[2]{Netguardians SA}

\begin{document}

\maketitle

\begin{abstract}

In classification problems, supervised machine-learning methods outperform traditional algorithms, thanks to the ability of neural networks to learn complex patterns. However, in two-class classification tasks like anomaly or fraud detection, unsupervised methods could do even better, because their prediction is not limited to previously learned types of anomalies. An intuitive approach of anomaly detection can be based on the distances from the centers of mass of the two respective classes. Autoencoders, although trained without supervision, can also detect anomalies: considering the center of mass of the normal points, reconstructions have now radii, with largest radii most likely indicating anomalous points. Of course, radii-based classification were already possible without interposing an autoencoder. In any space, radial classification can be operated, to some extent. In order to outperform it, we proceed to radial deformations of data (i.e. centric compression or expansions of axes) and autoencoder training. Any autoencoder that makes use of a data center is here baptized a centric autoencoder (cAE). A special type is the cAE trained with a uniformly compressed dataset, named the centripetal autoencoder (cpAE). The new concept is studied here in relation with a schematic artificial dataset, and the derived methods show consistent score improvements. But tested on real banking data, our radial deformation supervised algorithms alone still perform better that cAEs, as expected from most supervised methods; nonetheless, in hybrid approaches, cAEs can be combined with a radial deformation of space, improving its classification score. We expect that centric autoencoders will become irreplaceable objects in anomaly live detection based on geometry, thanks to their ability to stem naturally on geometrical algorithms and to their native capability of detecting unknown anomaly types.
\end{abstract}

\newpage

\section{Introduction}
Anomaly detection, also known as outlier detection, is the process of detecting data values that deviate significantly from the majority of the data. Anomaly detection has been an active research area for several decades, and it has a wide variety of applications in specific domains, including fraud detection and network intrusion detection as well as in broader domains such as risk management, AI safety, medical risks. Most standard classifiers such as decision trees assume that learning samples are evenly distributed among different classes. However, in many real-world applications, the propotion of the minority class is very small (can exceed one over 1 million). Due to the lack of data, few samples of the minority learning class tend to be falsely detected by the classifiers and the decision limit is therefore far from accurate. Research works in machine learning has been proposed to solve the problem of data imbalance (\cite{Garcia}, \cite{Galar}, \cite{Kraw}, \cite{Abra}, etc). However, most of these algorithms suffer from certain limitations in real-world applications, such as the loss of usual information, classification cost, excessive time, and adjustments, see (\cite{Abra}). There are also other useful tools to detect outliers, such as Principal Component Analysis (PCA) (\cite{Huang2006} \cite{chen2017detection}) but they use linear transformations. In contrast autoencoders, a special case of neural networks, with non-linear dimensionality reduction capability (\cite{hwang1999,hinton2006reducing}), use transformations with non-linear activation functions and multiple layers. They have been used for 30 years (\cite{grubbs1969}) and it has been shown that autoencoders are able to detect subtle outliers. PCA cannot (\cite{sakurada}). Many papers favor the use of autoencoders for dimensionality reduction (\cite{hinton2006reducing}) over PCA, not only because of its expressivnesss but also because it is more efficient to learn many layers with it rather than learn a huge transformation with PCA.  As we said, autoencoders are used to reduce the size of our inputs into a smaller representation. When we need the original data, we can reconstruct it from the compressed data. Outliers and anomalies can be detected by differenciating the input and the output of the autoencoder. More broadly autoencoders are also used for image coloring, feature variations, dimensionality reduction or denoise images. 

\section{Related Work}

There are many survey papers, tutorials and books (\cite{chandola2009anomaly}, \cite{aggarwal}) which describe the area of anomaly detection in a fairly broad way. Much work has also been done in the direction of autoencoders for outliers detection, e.g. \cite{roy2019robust,finke2021,Martinelli2004,schirrmeister2020,golan2018deep} and interestingly, there has been a recent resurgence of interest in developing unsupervised methods for anomaly detection lately (\cite{ruff2021unifying}, \cite{pang2021deep}). Miscellaneous forms of deep autoencoders have been used for reconstruction-based anomaly estimation. In (\cite{xia2015learning}) they study the problem of automatically removing outliers from noisy data by using reconstruction errors of an autoencoder. A variational autoencoder was used by An and Cho (\cite{an2015variational}), to estimate the reconstruction probability by Monte Carlo sampling, from which the outlier score is extracted. Another related method, which evaluates an unseen sample based on the model's ability to generate a similar one, uses generative adversarial networks (GANS) (\cite{goodfellow2014generative}). However most of the work concern anomalies in pictures in various domains. Our problem is  to find anomalies in timeseries (\cite{moschini2021anomaly}, \cite{houssou2019adaptive2}), which is an important research topic with many applications in the industry for example, to help monitor key metrics and alert for potential problems (\cite{ren2019time}).  To address theses issues LSTM-based encoder-decoder have been proposed (\cite{malhotra2016lstm}) to reconstruct normal probabilistic behaviors of timeseries. The reconstruction errors are used to detect anomalies. The same type of technique has been used for telemetry data (temperature, radiation, power, instrumentation, computational activities,...) in spacecraft missions (\cite{hundman2018detecting}). OmniAnomaly (\cite{su2019robust}) also proposes a stochastic recurrent neural network for time series anomaly detection but to the best of our knowledge, none of previous works in the literature have addressed the new type of ideas outlined in this paper.


\section{Motivation of radial methods}
Any geometric space endowed with a distance can underlie intuitive classifications. When dealing with a two-class imbalanced dataset, the distance from the center of the most populous class (gathering the normal points) can naturally serve as the only criterion when a point needs to be described as normal or anomalous. This simple criterion describes a \textbf{radial baseline} for any classification that follows; it is usually synthesized in one number or classification score (e.g. the area under the ROC curve). In datasets, axes have different meanings and their respective numbers should be treated distinctively. One can give them different weights, which corresponds to a geometrical transformation of the space. Because we keep the dataset center constant, these operations are named \textbf{radial deformations}. One can maximize the classification performance by optimizing the axis weights, referred in this document as \textbf{feature factors}.  Adding autoencoders (AEs) in the spectrum of geometrical transformations stems from two properties: First, they preserve the input dimension in their output, as radial deformations do. Second, they are non-linear functions, as opposed to radial deformations, and that can bring more to the table. Like in he case of radial deformations, classification with the presented AEs makes use of the radii of the output points with respect to the center of the dataset. This is why we call them \textbf{centric autoencoders}. We note from the beginning that classification methods are supervised or not. While radial deformations belong to the spectrum of supervised methods, AEs are based on unsupervised learning. Studying them together is justified by their shared ability to transform a geometrical space. Moreover, they can be combined in order to improve overall classification.


\section{Radial classification}
\label{sub:radial_class}
\begin{figure}
      \centering
      \includegraphics[width=0.7\textwidth]{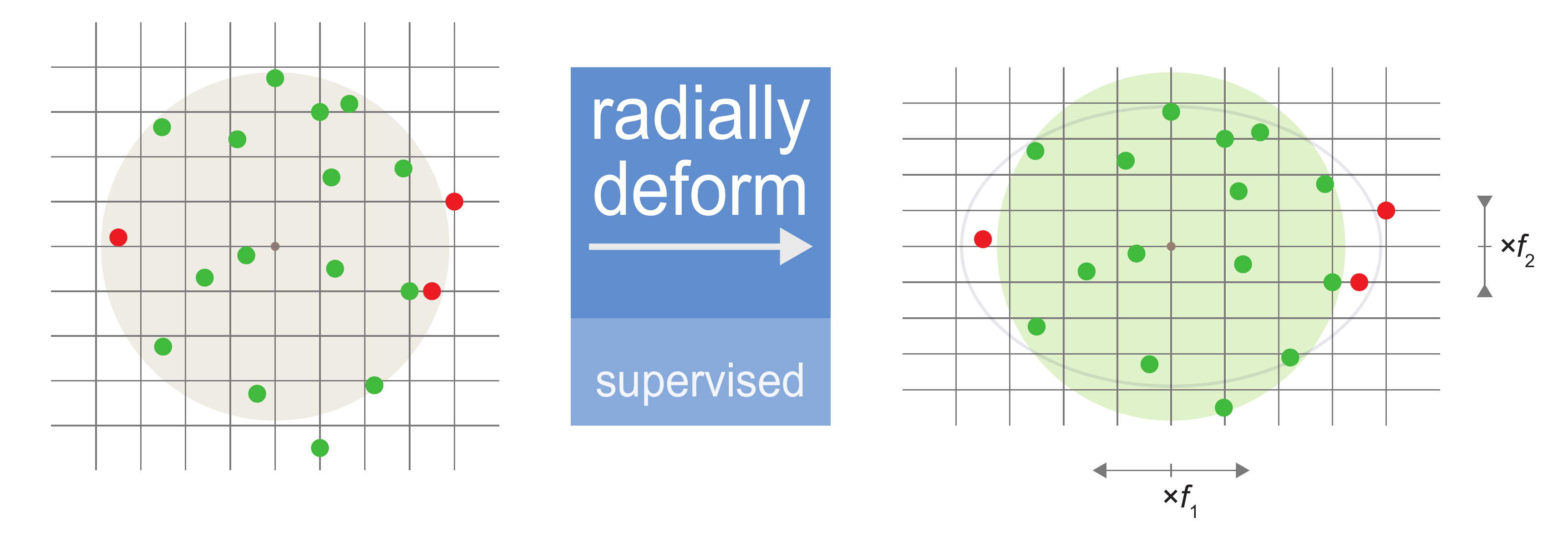}
      \caption{Radial deformation of space. In this two-dimension example, the horizontal direction is expanded, the vertical one is compressed about the center ($f_1>1$, $f_2 < 1$). This way, the red points (anomalies) are pushed away, while the lowest green point is brought closer to the center.}
      \label{fig:radial}
    \end{figure}

Within a distance-endowed space, one can define a center for the dataset, then sort its points based on their radii. In a two-class problem, points fall either in one class if their radii is lower than an established threshold or in the second class otherwise. The choice of the center point aims at favoring one class over the other. For instance, the center of mass of one first class will ensure likely lower radii for the points of the class itself. However, the center can be any point of the space, depending on the dataset structure and the learning objective. This study deals with two-class imbalanced datasets and employs the mass center of the predominant class. In general, functions can be applied on the dataset in order to obtain a new radii distribution leading to a better classification. In such a case, one investigates the positions of the function outputs with respect to the dataset center. Our targeted methods---radial (linear) deformations and autoencoders---are such functions. In the case of anomaly detection, the hypothesis is that suspicious points have different spreads on different axes of space, such that many axes present a larger standard deviation of the normal points. Hence, the radial baseline (a radius threshold suggested by the gray disk in fig.~\ref{fig:radial}) does not separate well between classes. Resizing the axes with different factors around a center may allow normal points to have smaller spreads on all axes---and enhances the radial classification (see the green disk in fig.~\ref{fig:radial}).
	
    \begin{definition}
      A \textbf{radial deformation} of a dataset $\mathcal D$ is a transform that compresses or expands the axes around their respective centers: 
      \begin{equation}
         x_i \rightarrow {x_C}_i + f_i \times (x_i - {x_C}_i) \;\; \forall x \in \mathcal D,
      \end{equation}
      with $i$ the axis number, $x_C = \mathbb E [\mathcal D]$ the center of the dataset, and $f_i$ the radial deformation \textbf{factor} of the $i$-th feature. 
    \end{definition}

    \subsection{Algorithms for radial deformation}
    To optimize classification via radial deformations, we developed two methods: - a greedy search of factors and the - angular gradient ascent on the factor hypersphere.
   
   \begin{itemize}
   \item  \textbf{Greedy search} optimization: Imagine first a simple and inexpensive way of optimizing a trained cAE: For each particular feature, we inflate or compress only the feature values of the dataset by a positive number (the deformation factor), either lower than 1 (inflation) or greater than 1 (compression); Calculate the classification score (e.g. AUROC) and repeat for different factors from a predefined range; Retain the one with the best score; Repeat for the other features in order to obtain their own deformation factors, too. At the end, a vector of radial factors is obtained. We empower this simple algorithm with the following enhancements: A) When starting to process the $i$-th axis ($i$-th feature), search for the magnitude order first, i.e. use a deformation factor $f_i$ like $10, 10^2, 10^3, ...$ and $10^{-1}, 10^{-2}, 10^{-3}, ...$ until the classification score for the entire dataset does not improve anymore. Once the best magnitude order is found to be $m$, calculate scores for $f_i = 1\cdot 10^m,\; 2\cdot 10^m,\; 3\cdot 10^m,\; ...$ until the score stagnates, and retain that digit (this would be $f_i$'s first non-zero digit, $d_0$). Then search for the second digit ($d_1$) by adding to $f_i$ the quantities $0, \; 1\cdot 10^{m-1},\; 2\cdot 10^{m-1},\; 3\cdot 10^{m-1}, ...\;$, then compute the third digit ($d_2$) in a similar way, and so on. The chosen factor will be $f_i = d_0 \cdot 10^{m} + d_1\cdot 10^{m-1} + d_2 \cdot 10^{m-2} + ... \;$; usually, three digits are largely enough.  Keeping this value for $f_i$, go on with processing the next axis, $i+1$. B) Another improvement is the acceptance of negative values for `digits`. Namely, $\exists j \; d_j<0$ and any digit $d_k$ is a negative or positive correction for the magnitude order $m-k$, i.e. $d_k \cdot 10^{m-k}$. Technically, one first looks for positive corrections ($d_k = 1, 2, ..., 9$), then for negative corrections ($d_k = -1, -2, ..., -9)$. This is a greedy algorithm, because it processes feature by feature and sequentially produces classification optima at the end of each axis computation. This does not guarantee that the final score is optimal. For this reason, we may see score enhancements when we run the algorithm several times (without resetting the factors). The detailed pseudo-code is presented in algorithm \ref{alg:greedy}).
      
      \begin{algorithm}
        \caption{Search of deformation factors that individually optimize radial classification \label{alg:greedy}}
        \begin{algorithmic}[1]
          \Require{the training set, $\mathcal D$; the scoring function, $s$}; how many digits, $d$, should be in a factor.
          \Statex We commonly use $s = \textrm{AUROC}$. The number of digits, $d$, does not include the leading zeros (e.g. the factor $f=0.00085$ complies to the requirement $d=2$ digits).
          \Function{ComputeFactors}{$\mathcal D, s, d$}
            \Let{$N\sub{axes}$} {number of axes (features) in $\mathcal D$}
            \Let{$score$} {$s(\mathcal D)$}
            \Let{$f$} {$[1, 1, ..., 1]$} \Comment{factors: an array with $N\sub{axes}$ elements}

            \For{$i \gets 1 \textrm{ to } N\sub{axes}$} \Comment{calculate $f_i$}
                \Let{$order$} {0} \Comment{start computing the order of $f_i$}
                \Repeat
                	\Let {$order$} {$order + 1$}
                	\Lets {$f'$} {$f$} {$f'[i]$} {$10^{order}$}
                	\State $\mathcal D'$ = \Call{Expand}{$\mathcal D, f'$} \Comment{deforms the data with the centric factors $f'$}
                	\Let {$new\_score$} {$s(\mathcal D')$}
                	\If {$new\_score \leq score$}
                		\Let {$score\_stagnates$}{True}
                		\Let {$order$} {$order - 1$}
							\Else 
								\Let {$score\_stagnates$}{False}
								\Let {$score$} {$new\_score$}
							\EndIf
						\Until {$score\_stagnates$}

						\If {$order = 0$} \Comment{no positive orders enhanced scores}
							\Repeat \Comment{hence try negative orders}
                		\Let {$order$} {$order - 1$}
                	\Lets {$f'$} {$f$} {$f'[i]$} {$10^{order}$}
                	\State $\mathcal D'$ = \Call{Expand}{$\mathcal D, f'$} 
                		\Let {$new\_score$} {$s(\mathcal D')$}
                		\If {$new\_score \leq score$}
                			\Let {$score\_stagnates$}{True}
                			\Let {$order$} {$order + 1$}
								\Else 
									\Let {$score\_stagnates$}{False}
									\Let {$score$} {$new\_score$}
								\EndIf
							\Until {$score\_stagnates$}
						\EndIf

						\Let{$f[i]$} {$10^{order}$} \Comment{write down the first computation of $f_i$}
						\State ... \Comment{here comes code that determines}
						\State ... \Comment{the first non-zero digit of $f_i$}
						
						\For{$q \gets 2 \textrm{ to } d$} \Comment{compute the next digit for  $f_i$}
							\Let {$order$} {$order - 1$}
							
							\Let {$positives$} {$[1, 2, 3, ..., 9]$}
							\Let {$negatives$} {$[-1, -2, ..., -9]$} \Comment{negative digits are negative corrections}
							\For {$range \gets positives \textrm{ or } negatives$}
							
								\For {$digit \textrm{ from } range$}
                			\Let {$f'$} {$f$}
                			\Let {$f'[i]$} {$f[i] + digit \times 10^{order}$}
                			\State $\mathcal D'$ = \Call{Expand}{$\mathcal D, f'$}
                			\Let {$new\_score$} {$s(\mathcal D')$}
                			\If {$new\_score \leq score$} \Comment{the score stagnates}
                				\State \textbf{break}
									\Else \Comment{the score enhanced}
										\Let {$score$} {$new\_score$}  \Comment{update the score}
                				\Let {$f[i]$} {$f'[i]$}  \Comment{update $f_i$}
									\EndIf
								\EndFor

							\EndFor
						\EndFor
            \EndFor
            \Return {$f, score$}
          \EndFunction
        \end{algorithmic}
      \end{algorithm}

 \item    \textbf{Angular gradient ascent} on the factor hyperspere. All factors form a $N$-dimensional vector, that we call the factor vector. The starting point in a radial deformation is a vector of $N$ equal factors. Note that, given any combination of factors, multiplying all factors by the same number does not produce any change in the relative positions of points. Hence, some factor vector $\vec{v}$ of norm 1 is equivalent to any multiple of $\vec v$. Therefore, searching better factors can be restrained on the surface of an $N$-dimensional hypersphere (also known as the $N$--1-sphere) of radius 1. For convenience, one may choose to operate only on the equivalent  hypersphere of radius $\sqrt N$, because the undeformed space has the factor vector $(1, 1, ..., 1)$, of norm $\sqrt N$. The hypersphere is described by $N-1$ angles. Our sphere walk function implements the established method of gradient descent/ascent, with the specific aspect that here coordinates are angles. We also developed a stochastic version of the function, that reduces execution times on big datasets.

\end{itemize}

\section{Building the concept of centric autoencoder}
\label{sec:concept_centric_ae}

  The existential purpose of autoencoders used to be object compression and reconstruction. We re-purpose them for object classification, by making use of the Euclidean distance in the output space. 
  Our interest is to build autoencoders whose outputs have a better radial separation than the dataset points (the radial baseline) and to come closer to the performance of the supervised methods of radial classification. For completion, comparisons with vanilla (non-centric) AEs will be added. The exercise of combining AEs with radial classification starts with the next definition.

    \begin{definition}
      A \textbf{centric autoencoder} (cAE) is an autoencoder with a defined center, that classifies its inputs by their respective outputs' radii from the center. 
    \end{definition}
      
    The following subsections study some relevant flavors of centric autoencoders.
  \subsection{The cAE: any AE radii-based classifier}
    \label{sub:bare}
\begin{figure}
      \centering
      \includegraphics[width=0.7\textwidth]{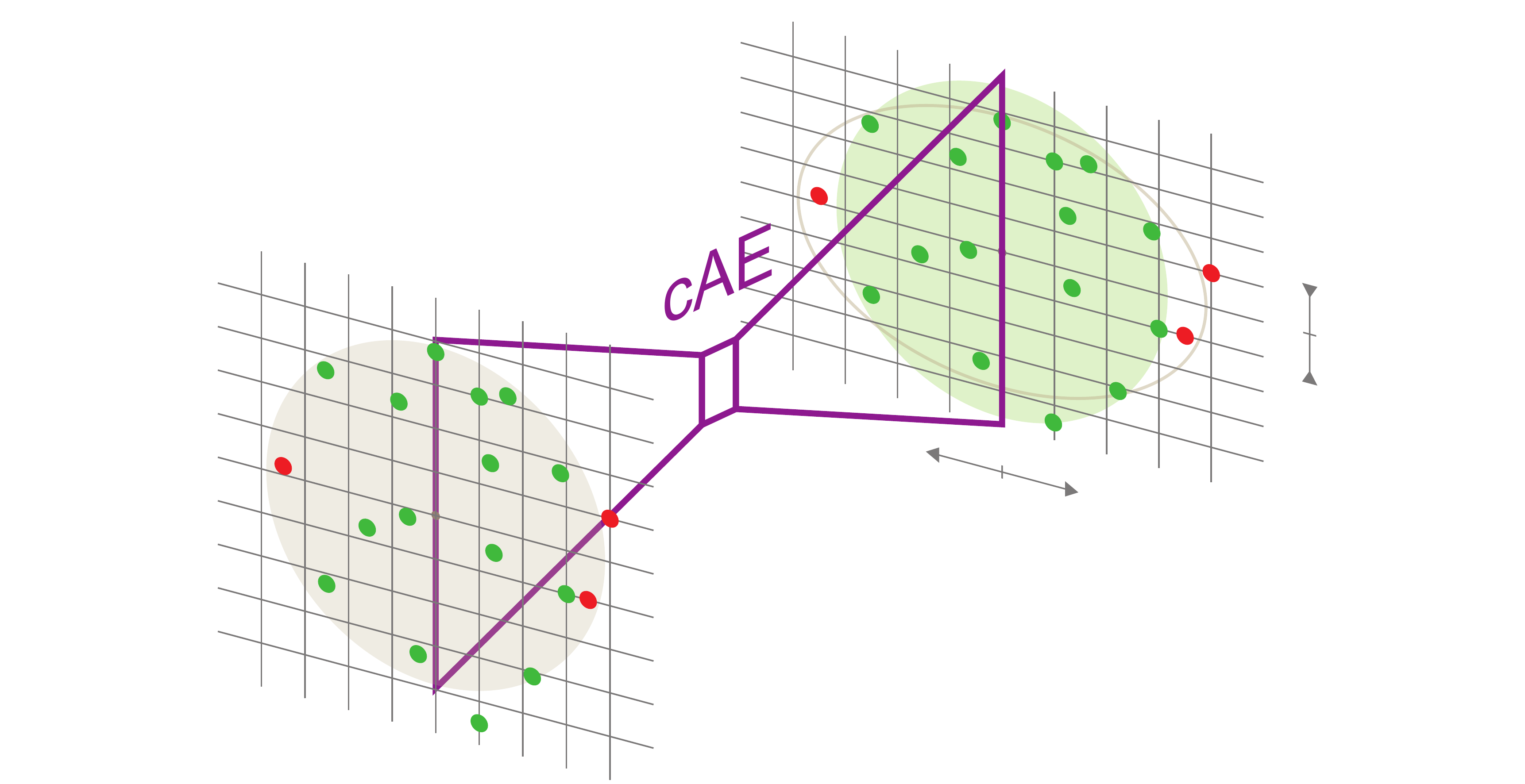}
      \caption{A centric autoencoder aims at transforming space, if possible in the same way as a radial deformation. Here, the red points (anomalies) are pushed away, while the lowest green point is brought closer to the center.}
      \label{fig:cae}
    \end{figure}

    An autoencoder trained for input reconstruction, if given a center (e.g. the mass center of one of the classes in the training set), can be used as a centric autoencoder. Fig.~\ref{fig:cae} presents a general cAE that uses the dataset center. The obtained space transformation is here equivalent to a radial deformation. Let us explore if a cAE can produce this kind of effects.

\begin{figure}
      \centering
      \includegraphics[width=0.9\textwidth]{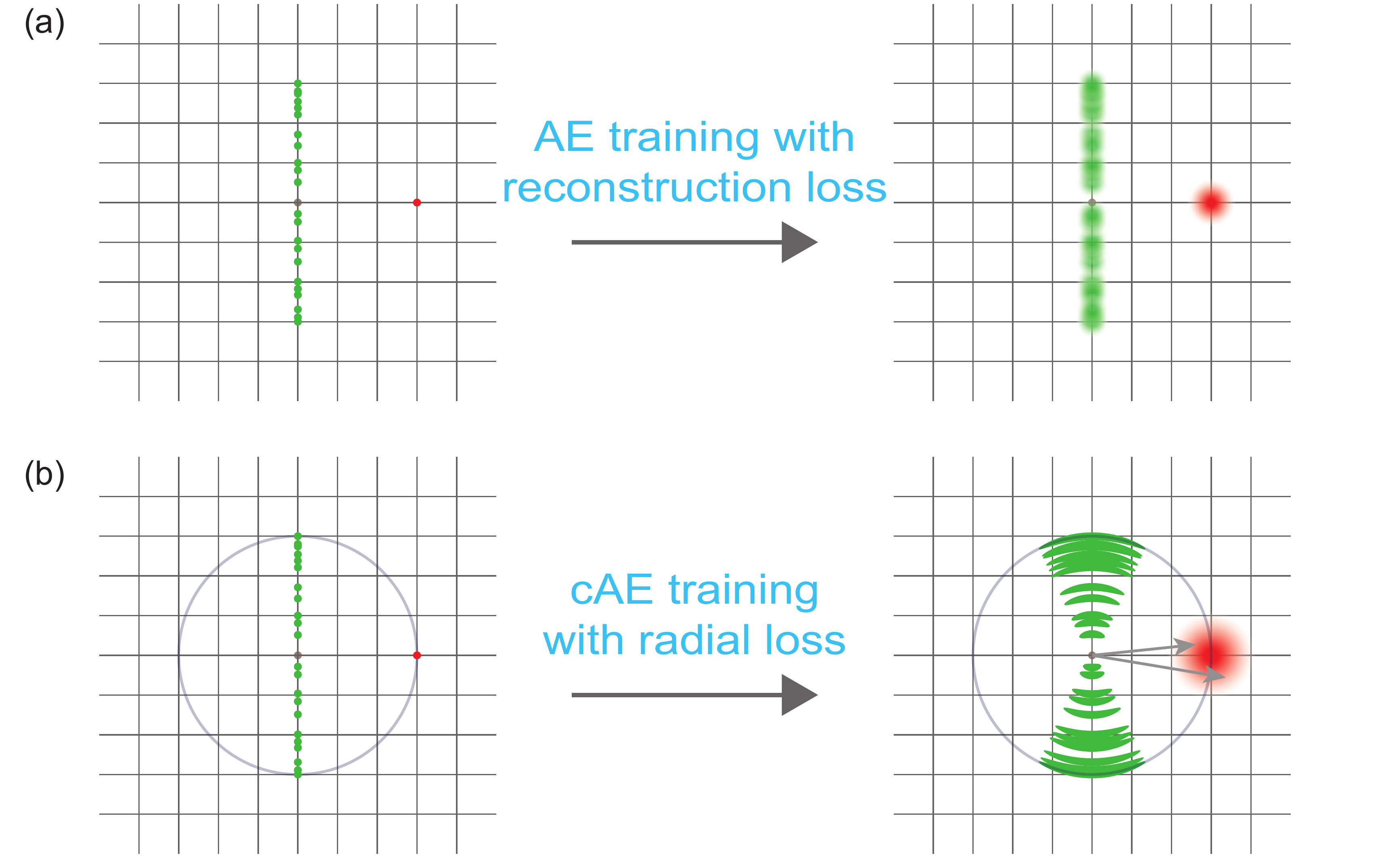}
      \caption{(a) The training of a general AE tries to minimize reconstruction errors. (b) When a cAE is trained by punishing radius change, anomalous points have an higher chance of radius increase: the red region is more outside the circle than inside.}
      \label{fig:cae_loss}
    \end{figure}

\textbf{Can a loss function deform space?} For this question, we consider an input 2D-space where normal points have a rather small spread on the horizontal axis, while anomalies deviate little on the vertical axis (fig.~\ref{fig:cae_loss}). In the first scenario (fig.~\ref{fig:cae_loss} a), we choose AEs' traditional loss function, i.e. the cAE is trained by minimizing the reproduction error. Although anomalies should show higher reconstruction errors (their region did not participate in training), the output space do not underlie a geometrical transformation similar to radial deformations. In the second scenario (fig.~\ref{fig:cae_loss} b), we conceive a function that punishes the radius change. This is the radius relative variation between input and output:

    \begin{equation}
    		\rho = \frac {|r\sub{out} - r\sub{in}|} {r\sub{in}}.
    \end{equation}
By minimizing the radial loss function, $\sum_i \rho_i$, points will be reconstructed on arcs that preserve the initial radii. Nonetheless, the anomaly reconstructed in the figure does not necessarily catch the same reconstruction pattern as normal points, because its sector was ignored in the training phase. If we consider its probable reconstructions do not really favor any particular direction (similarly to the Gaussian distribution illustrated in the right part of fig.~ \ref{fig:cae_loss} a), then their region extends more outside the traced circle (fig.~\ref{fig:cae_loss} b) than inside. That means the output radius increased, $r\sub{out} > r\sub{in}$. In other terms, the anomalous regions are pushed farther from the center, and a radial deformation occurs. In the following, only the radial loss function is used. In the same spirit, the loss function can force radius reduction for normal points, producing a swifter radial deformation. For example, $\rho = {|2 r\sub{out} - r\sub{in}|} / {r\sub{in}}$ will cut output radii of normal points by half. However, we did not apply radial-reduction loss functions in our project. In conclusion, the answer to the above question is yes, radial loss functions deform space. In general, expansion will occur on axes where anomalies have a larger spread than normal points have. Loss functions that embed radius reduction accentuate this effect. Under this light, the last plane of fig.~\ref{fig:cae} is correct in the sense of preferential expansion directions. As for compression in other directions, they are to be achieved with bias units or activation functions.

\begin{figure}
      \centering
      \includegraphics[width=0.9\textwidth]{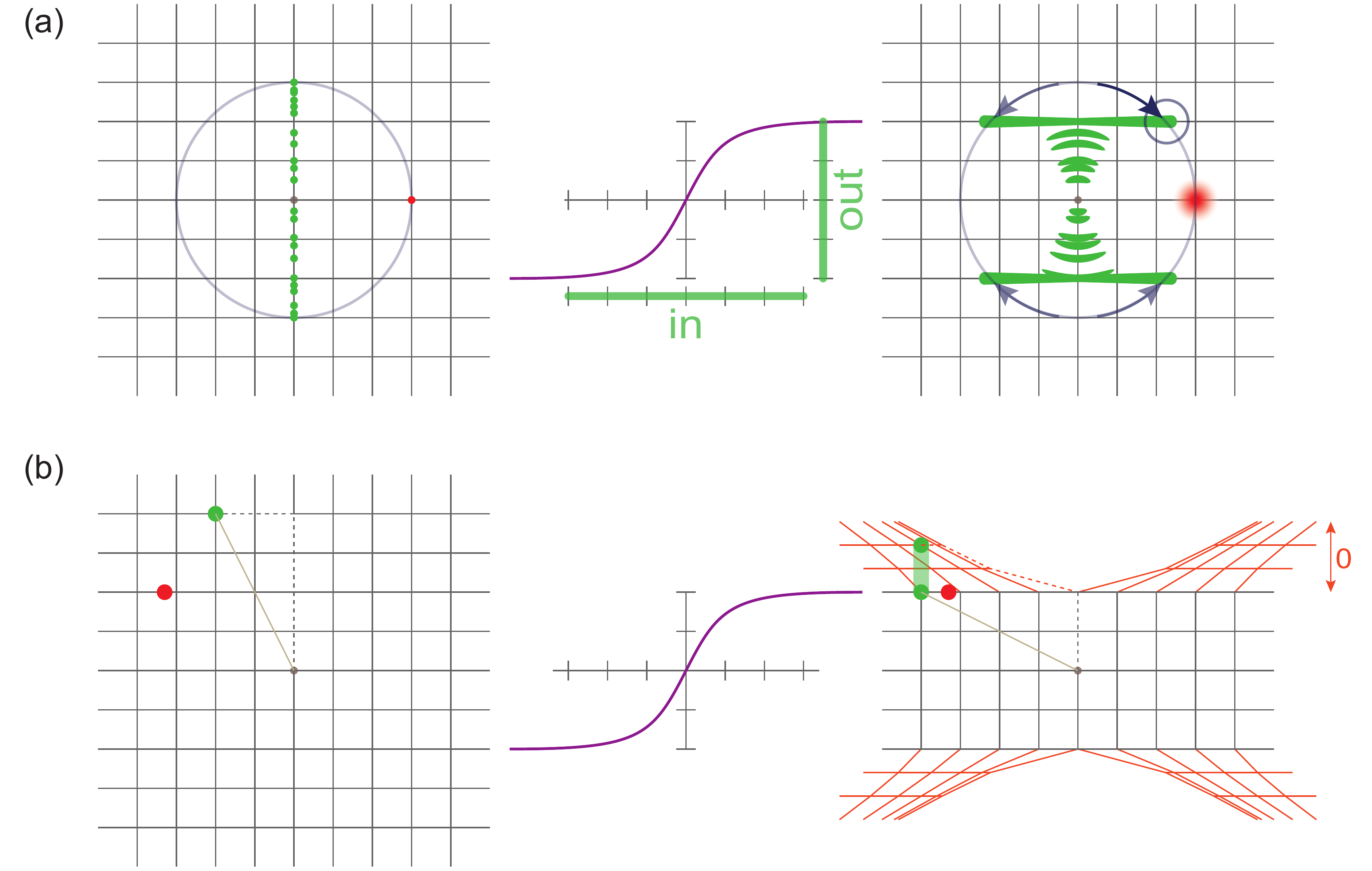}
      \caption{Consider an activation function for the vertical axis. (a) Due to activation, reconstruction with a radial-loss function can produce artifacts (\emph{antennas}) in the cAE output. Do their ends correspond to radial changes larger than those of anomalies? (b) How the space fabric changes in the above situation. The red mesh has no vertical width, but is artificially stretched for visualizing how the initial grid points migrated laterally. On the low-spread direction of the dataset (here, the horizontal axis), a normal point (green) is thrown beyond an anomalous one (red dot).}
      \label{fig:cae_antennas}
    \end{figure}

    \textbf{Artifacts in the realistic cAE.} Neural networks are more powerful when their architecture include bias units and activation functions. Let us suppose we have only one activation function, in the output layer (e.g. a softsign-like function for the vertical axis fig.~\ref{fig:cae_antennas} a). Note first this is not unusual for AEs. If you cannot accept that such a small-output activation function is employed intentionally, knowing that the dataset have larger values on the concerned axis, you can just suppose that data values arrive enlarged at the last layer. The activation function has a threshold that forces larger-value points to slide further on their reconstruction arc, until meeting the cap (follow arrows in fig.~\ref{fig:cae_antennas} a). Vertical shrinkage is compensated with horizontal elongation. The newly formed elongated artifacts are here called \textbf{compensation antennas}. Longer slides come in practice with less precise reconstructions on the circle. The question arises if these antennas present, at least at their ends, radial variations comparable to the those of anomaly reconstructions. The space deformation that corresponds to compensation antennas is illustrated in  fig.~\ref{fig:cae_antennas} b, where the deformed output fabric marked in red has no $y$-width. In fact, the red mesh is obtained by vertically stretching the upper and lower regions. This allows us to visualize how the initial grid points migrated laterally.  This type of radial deformation works against our purpose, as it extends space in the opposite direction to the desired one. One can also argue that the same phenomenon can occasionally be triggered by bias units, too. Later in this development, we present methods to counter this undesired effect.
	
    \textbf{Methodology detail.} During the training phase of a cAE, it is practical to watch for classification performance. While training adjusts the autoencoder weights to minimize reproduction errors (reproduction in the way defined above), we calculate after each training epoch the classification performance over a test set. This way, one can select from the training track those weights that generated the best classification over the test set, before reconstruction losses lower into an overfit regime. 
    
  \subsection{The centripetal autoencoder, cpAE: the training set is compressed}
    \label{sub:common}
\begin{figure}
      \centering
      \includegraphics[width=0.6\textwidth]{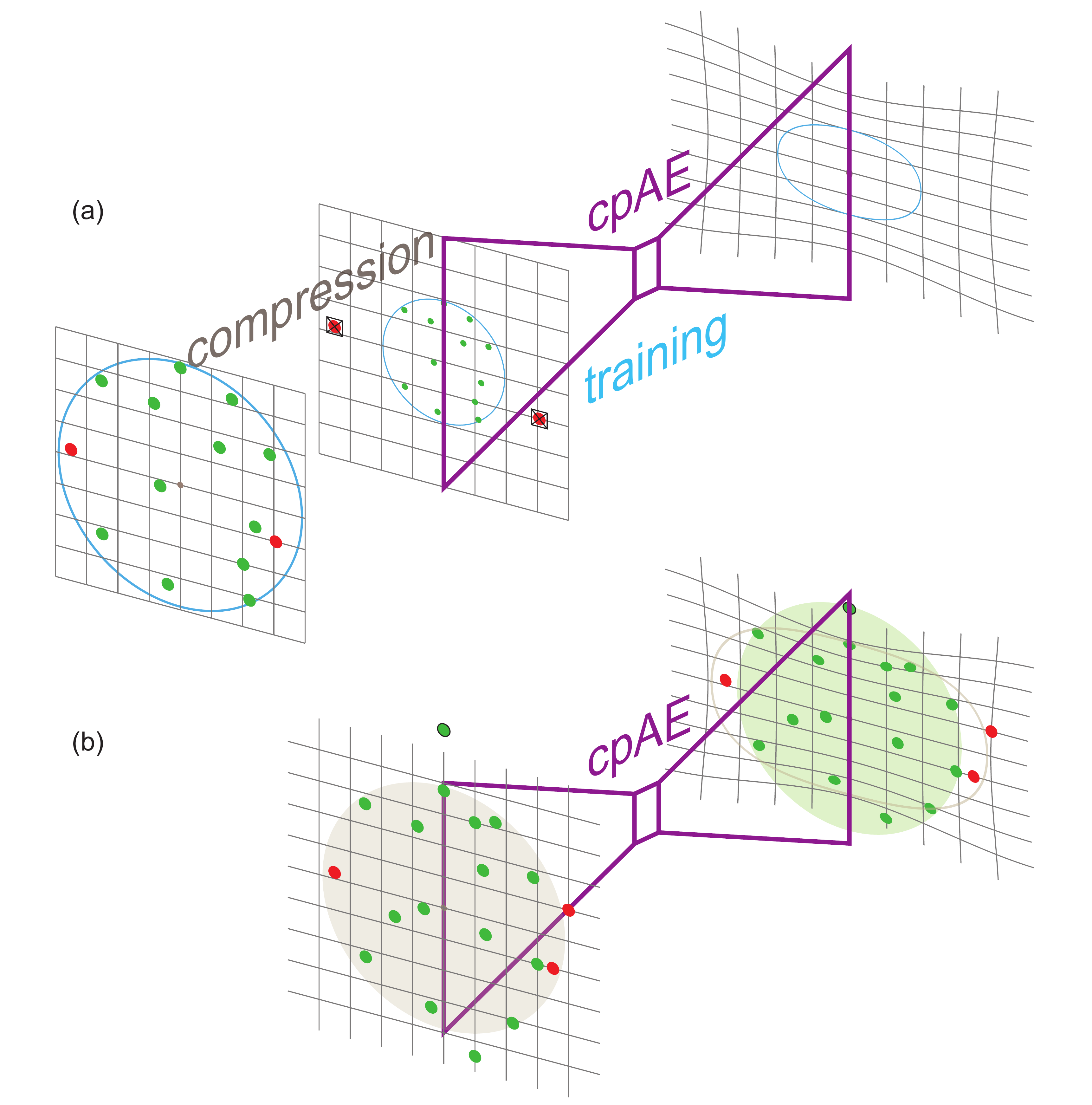}
      \caption{ The centripetal autoencoder, cpAE. (a) cpAE in the training phase. (b) cpAE in prediction mode. The gray-contour oval marks the limit of the (input) radial baseline. Remark that the input outlier (highest input point) is correctly classified at the output end.}
      \label{fig:cpae}
    \end{figure}

    \textbf{Intuitive motivation.} In an anomaly detection problem, we asked ourselves if an autoencoder strengthens the memorization (i.e. reproduction with lowest errors) of the most typical points when trained with more points close to the center of the dataset. Compressing the training set (which contains no anomalies) towards its center of mass exposes the autoencoder only to points similar to the statistical mean of the dataset. This is a form of mean-region memorization, at least for overparametrized autoencoders  (\cite{radhakrishnan2018}). During prediction, reproductions will be attracted by the memorized region, with untypical inputs being reconstructed farther away from the center. This process enables radii-based classification. But if the number of neurons in the AE is relatively low, then we cannot claim overparametrization. However, there could still be a lighter form of memorization: \emph{pseudo-memorization}. We can hope we deal with some reminiscent form of attraction, which is a non-linear radial deformation. Fig.~\ref{fig:cpae} presents a cpAE in the training phase (a), then in prediction mode (b). On the last plane of fig.~\ref{fig:cpae} a, one can see the deformed output fabric, with the locus of the training set delimited by an oval contour. Inside the locus, space very little deformed, like in the conceptual drawing of fig.~\ref{fig:cae}. Outside, this locus exercises attraction forces on the rest of the plane, and these forces are stronger in the vertical direction. Their effect is the input outlier (highest input point) being brought closer to the center in the output space.
	
    \begin{definition}
      A centric autoencoder of center $x_C$ trained with a compression $c\mathcal{T}$ of a dataset subset $\mathcal{T} \subset \mathbb{R} ^n$,
      \begin{equation}
        c\mathcal T \equiv  \{ x_C + c(x - x_C) | x \in \mathcal{T} \},
      \end{equation}
      is a \textbf{centripetal autoencoder} if $x_C$ is the center of mass of $\mathcal{T}$, $x_C = \mathbb E [\mathcal T]$. The centric factor $c$ is a scalar in $(0, 1)$.
    \end{definition}

    During training, the classification performance is to be calculated for an uncompressed test set. No data compression occurs at prediction time.
    
	\textbf{Deeper intuition for realistic cAEs.} Suppose all the data are centered in 0. In any AE without bias units and only ReLU (simple or leaky) activation functions, an isotropic radial deformation of factor $f$ applied on the input propagates along the neuron network up to the output. This statement is relatively easy to demonstrate. We call the \textbf{zero-bias limit} the situation where an AE has no bias units and only (simple or leaky) ReLU activation functions, except for the output layer. As for the above property of isotropic radial deformation propagation along an AE's layers, we name it the \textbf{zero-bias limit equivalence}. Given the fact that a cAE with an activation function in the output layer presents compensation antennas from higher values, the solution becomes clear: In the zero-bias limit, isotropic compression of inputs diminishes the outputs accordingly, and adding an activation function for outputs will not lead to artifacts like compensation antennas. In conclusion, by using a zero-bias cpAE instead of a cAE, one avoids counter-productive artifacts.  Did compensation antennas disappear in the cpAE prediction phase, too? In fact, compensation is coded in AE's weights when values hit the activation threshold, while training. As this did not happen anymore, antennas were lost permanently during training.
	
  \subsection{The expanded-input cAE}
    \label{sub:expanded}
    
\begin{figure}
      \centering
      \includegraphics[width=0.6\textwidth]{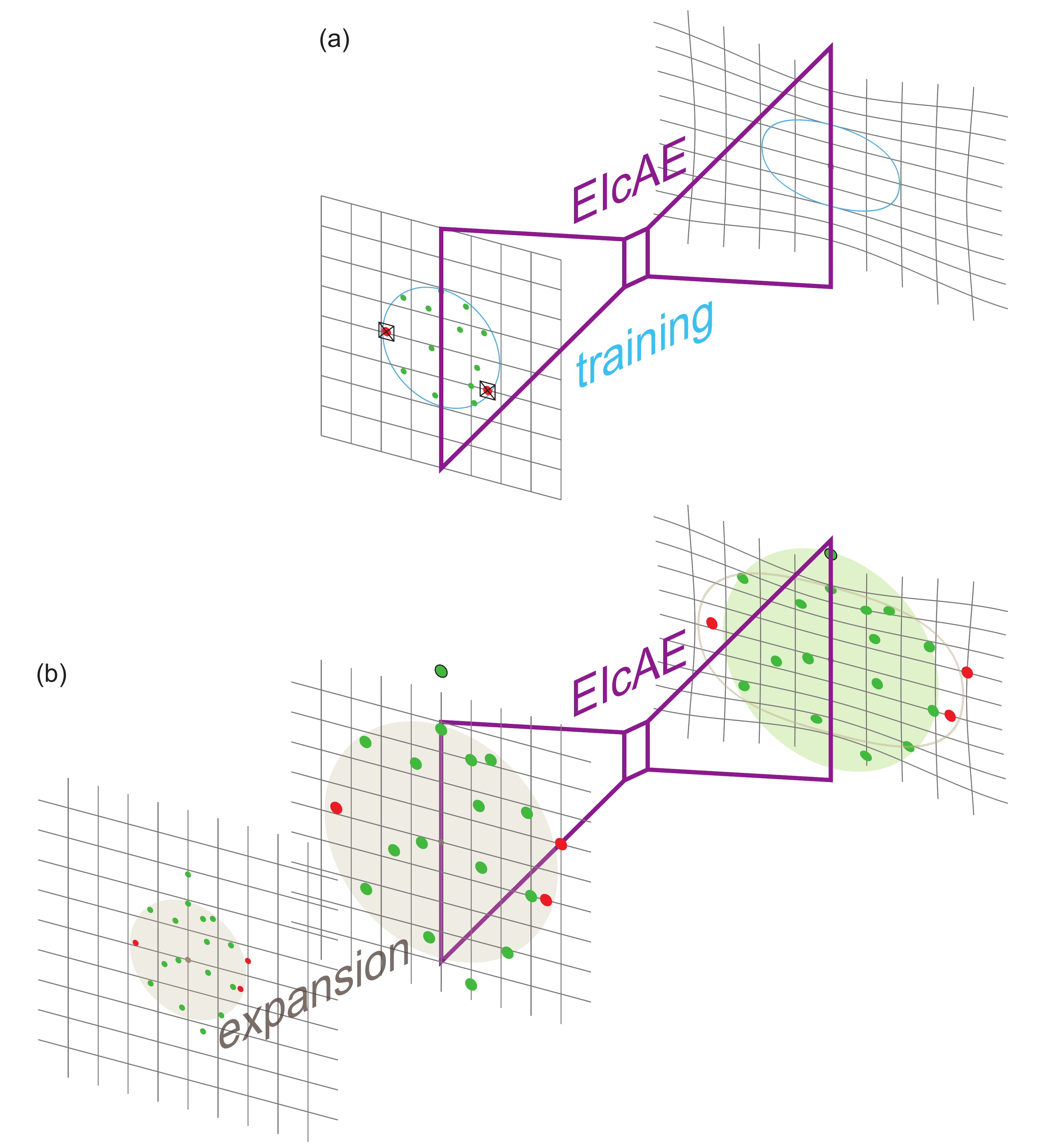}
      \caption{Extended-input centric autoencoder. (a) The training phase: a simple cAE. There is no effective activation function in the output layer, meaning no artifacts deform the output space. (b) In prediction mode, inputs are first expanded.}
      \label{fig:eicae}
    \end{figure}

    In this subsection, we first argue that the opposite of compression, expansion, can be used on inputs after training the cAE, with the same effect as the compression of inputs during training. This is nonetheless a limit case. We illustrate in fig.~\ref{fig:eicae} what an extended-input cAE is. Its hypothesis is that the output activation function, if any, has a limited effect, i.e. its threshold is large compared to the values in the dataset.
    
    The same way we talk about a training set of a neural network (NN), one can designate by the expression \emph{prediction set} all the input points that can be fed to the NN after its training. 
    
    \begin{definition}
      \label{reldist}
      The relative distance between the prediction set, $\mathcal P$, and the training set, $\mathcal T$, of a cAE of center $x_C$ is the ratio of the average radii in the two sets:
      \newcommand*\norm[1]{\Vert #1 \Vert}
      \begin{equation}
        \mathcal{PT} = \frac{\mathbb E \norm{x-x_C}_{x\in \mathcal P}} {\mathbb E \norm{x-x_C}_{x\in \mathcal T}}.
      \end{equation}
    \end{definition}
    Compressing a training set ($\mathcal T \rightarrow \check{\mathcal{T}}$) obviously increases the relative distance between the prediction and the training set (${\mathcal{PT}} < P\check{\mathcal{T}}$). Nonetheless, the same relative-distance effect occurs when the prediction set is expanded instead.

\begin{theorem}
\label{lemma}
The compression process involved in building a centripetal cAE is equivalent to input expansion during prediction, in the zero-bias limit.
\end{theorem}

The proof seems immediate: Apply definition \ref{reldist} to observe that $\mathcal P \check{\mathcal{T} }= \hat{\mathcal{P}} {\mathcal T}$, where $\check{\mathcal{T}} = c {\mathcal T}$ is the training set ${\mathcal T}$ compressed by a centric factor $c \in (0, 1)$ and $\hat{\mathcal{P}} = c^{-1} {\mathcal P}$ is the prediction set $\mathcal P$ expanded by a centric factor $c^{-1} = 1/c > 1$. Nonetheless, this proof ignores the existence of bias units in the layers of the NN and of an activation function its last layer. 

    \begin{definition}
      A trained centric autoencoder of center $x_C$ is an \textbf{expanded-input cAE} if, before prediction, it radially expands its inputs by a scalar expansion factor $e \in (1, \infty)$:
      \begin{equation}
        x \rightarrow x_C + e(x - x_C).
      \end{equation}
    \end{definition}
    At this point, we have found a quasi-alternative to the centripetal AE: using any trained cAE with expanded inputs. Instead of baptizing this alternative a centrifugal autoencoder, let us remark that prediction set expansion is complementary to training set compression, as they happen after and before training. Therefore, one can try tuning a cAE's performance by expanding inputs.

    \textbf{Intuition via a thought experiment.} In the zero-bias limit, let us analyze more the training of a cAE. Think that initially our numbers are expressed in a small measure unit (for instance, meters) and just train the cAE. Then we would like to use a large measure unit, like kilometers instead of meters. In km, numbers will become smaller: the dataset in km will have all values divided by 1000. If there is no activation function in the output layer, then we obtain the same results in the spirit of the zero-limit equivalence: cAE(m) = cAE(km). This is also what  intuition tells us about dealing twice with conceptually identical objects. But going from m to km was the compression phase of cpAE training: cAE$_t$(km) = cpAE$_t$(m) ($t$ for training). If there is some activation function in the output layer, then we realize that the measure unit of the activation output was not changed. In meters, it stayed a 1000x higher number, therefore the activation threshold was not reached anymore---exactly as in the illustrated cpAE (fig.~\ref{fig:cpae} a) that avoided this way compensation antennas. From here, one can go to prediction, whose story has been already told: go back to meters (higher numbers) and benefit from an artifact-free cAE. Continuing the thought experiment around measure units, let's take the opposite direction: Start now with a cAE trained in km. Then transition to meters (numbers become higher) and train a new cAE. In the zero-bias limit, the two are equivalent if the output layer has no activation function. Having an activation function in the last layer changes the effect of its numerically constant threshold: in the meter-picture, the threshold is heavily hit. Therefore, expanding numbers by a factor of 1000 before training should accentuate compensation antennas. What about transitioning from km to m only after training? The zero-bias limit with a passing-through activation function in the output layer is the above main case, covered by lemma~\ref{lemma}. With the trained cAE(km), predicting on 1000x inputs will hit the now hard-coded threshold, many points being now projected on the compensation antennas, which means lower performances. We are now left outside the zero-bias limit: When bias units are present, large inputs will act as bias neutralizers. In a biased world, compensation antennas are built with bias weights. Hence, expanded inputs (EI) in prediction can destroy antennas, annihilating the activation function. We find therefore a possible EIcAE-cpAE equivalence in this finite-bias space. We conclude that, in reality, many aspects are at stake when comparing different flavors of cAEs.
    
  \subsection{The deformed-output cAE}
    \label{sub:corrected}     
    After training a cAE, its output space (more precisely, the result of the cAE applied on the training set) can still be undercompressed on some axes, but overcompressed on other axes. Using supervised methods of radial deformation, the more general finding is that, after training, inputs should be recentered: expansively on some axes, compressively on the others.
    \begin{definition}
      \label{recentered}
      A \textbf{deformed-output cAE} is a cAE of center $x_C$ that is trained in two phases: the first one is the usual cAE training; the second one is a supervised radial deformation method applied on the first-phase output space.

    \end{definition}

\begin{figure}
      \centering
      \includegraphics[width=0.6\textwidth]{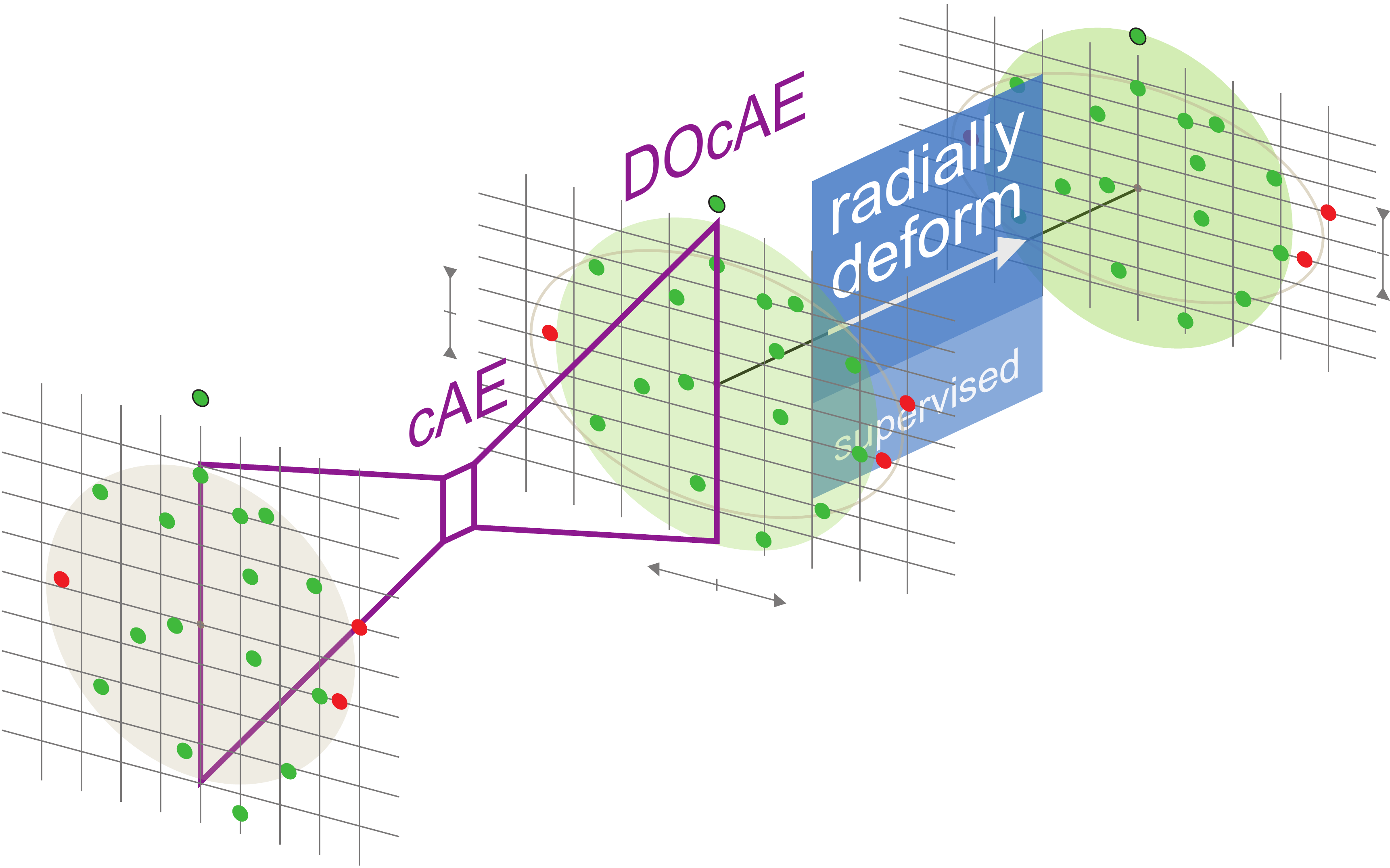}
      \caption{The deformed-output centric autoencoder is a hybrid object: a cAE (unsupervised method) and a radial deformer (supervised).}
      \label{fig:docae}
    \end{figure}

    Fig.~\ref{fig:docae} presents graphically the unsupervised-supervised hybrid concept of DOcAE. The aim of this hybrid approach is enhanced classification, hopefully up to the point it beats the geometrical radial deformation. We show that below can happen in practice, but it often underscores radial deformation.
     
\section{Experiment}
  We explore the properties of centric autoencoders while applying them on an artificial dataset. This dataset is constructed in a simple way and is enriched with statistic anomalies.
  \subsection{Constructed dataset}
    Our dataset has 15 features whose values are sampled from identical $\mathcal N (0, 1)$ normal distributions. 1\% of the points are altered in order to follow, on the first 12 axes out of the available 15, a narrower normal distribution, centered progressively farther from 0: $\mathcal N (i/16, 0.4), i=0..11$ (see algorithm \ref{alg:dataset}). The altered points are termed \emph{anomalies} or frauds, while the unaltered ones are \emph{regular} or legit points. The twelve axes on which anomalies have distinct distributions are called anomaly-revealing axes.

      \begin{algorithm}
        \caption{Creation of a simple dataset with anomalies \label{alg:dataset}}
        \begin{algorithmic}[1]
          \Require{the number of points, $N$; the ratio of anomalies, $r$}
          \Statex We commonly use $N=10^5$ and $r=0.01$.
          \Function{Dataset}{$N, r$}
            \Let{$f$}{$15$} \Comment{$f$: number of axes (features)}
            \Let{$f\sub{an}$}{$12$} \Comment{$f\sub{an}$: number of anomaly-revealing axes}
            \Let{$\Delta\mu$}{$1/16$} \Comment{$\Delta\mu$: step in anomaly distribution centers}
            \For{$i \gets 1 \textrm{ to } f$}
                \Let{$\mathcal D[i, 1:N]$}{\textbf{sample} from $\mathcal N(0, 1)$}
            \EndFor
            \For{$i \gets 1 \textrm{ to } f\sub{an}$}
                \Let{$\mathcal D[i, 1:rN]$}{\textbf{sample} from $\mathcal N \left( (i-1)\Delta\mu, 0.4\right)$}  \Comment{alter $rN$ values on axis $i$}
            \EndFor
            \Let{$y\sub{labels} [1:N]$}{$0$}
            \Let{$y\sub{labels} [1:rN]$}{$1$} \Comment{anomalies are labeled with $1$}
            \State \Return{$\mathcal D$}, $y\sub{labels}$
          \EndFunction
        \end{algorithmic}
      \end{algorithm}
      
\begin{figure}
      \centering
      \includegraphics[width=0.6\textwidth]{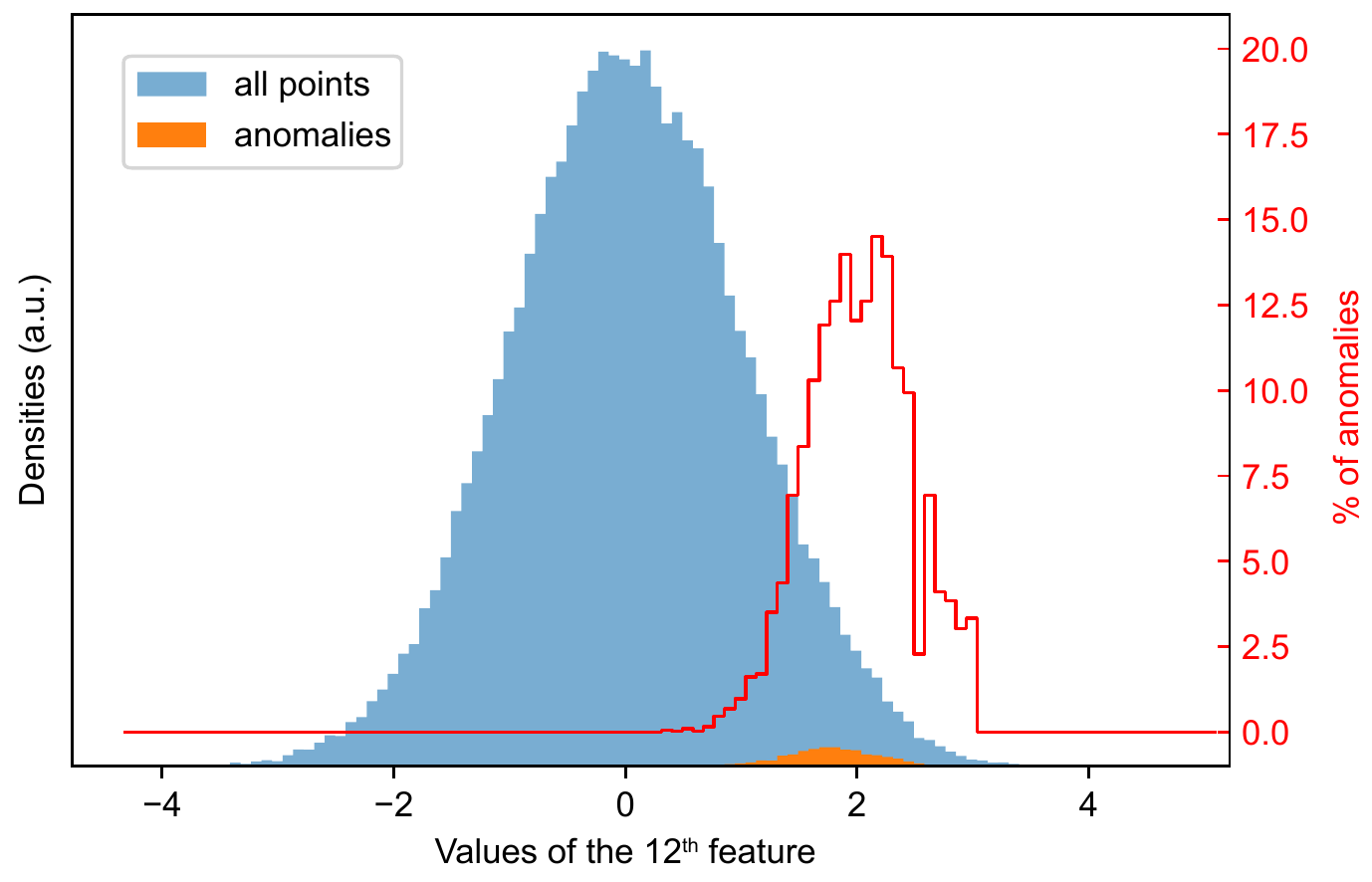}
      \caption{Histograms of the 12th feature for the entire dataset (blue) and anomalies (orange). The 12th axis is the one where anomalies are the most outcentered. Nonetheless, anomaly weight is lower than 15\% (red line).}
      \label{fig:distro11}
    \end{figure}

    When investigating the distribution of anomalies along any axis, one can see it is immersed into the distribution of all values, making anomalies apparently indistinguishable. Fig.~\ref{fig:distro11} presents these distributions for the most anomaly-revealing axis.

\begin{figure}
      \centering
      \includegraphics[width=0.6\textwidth]{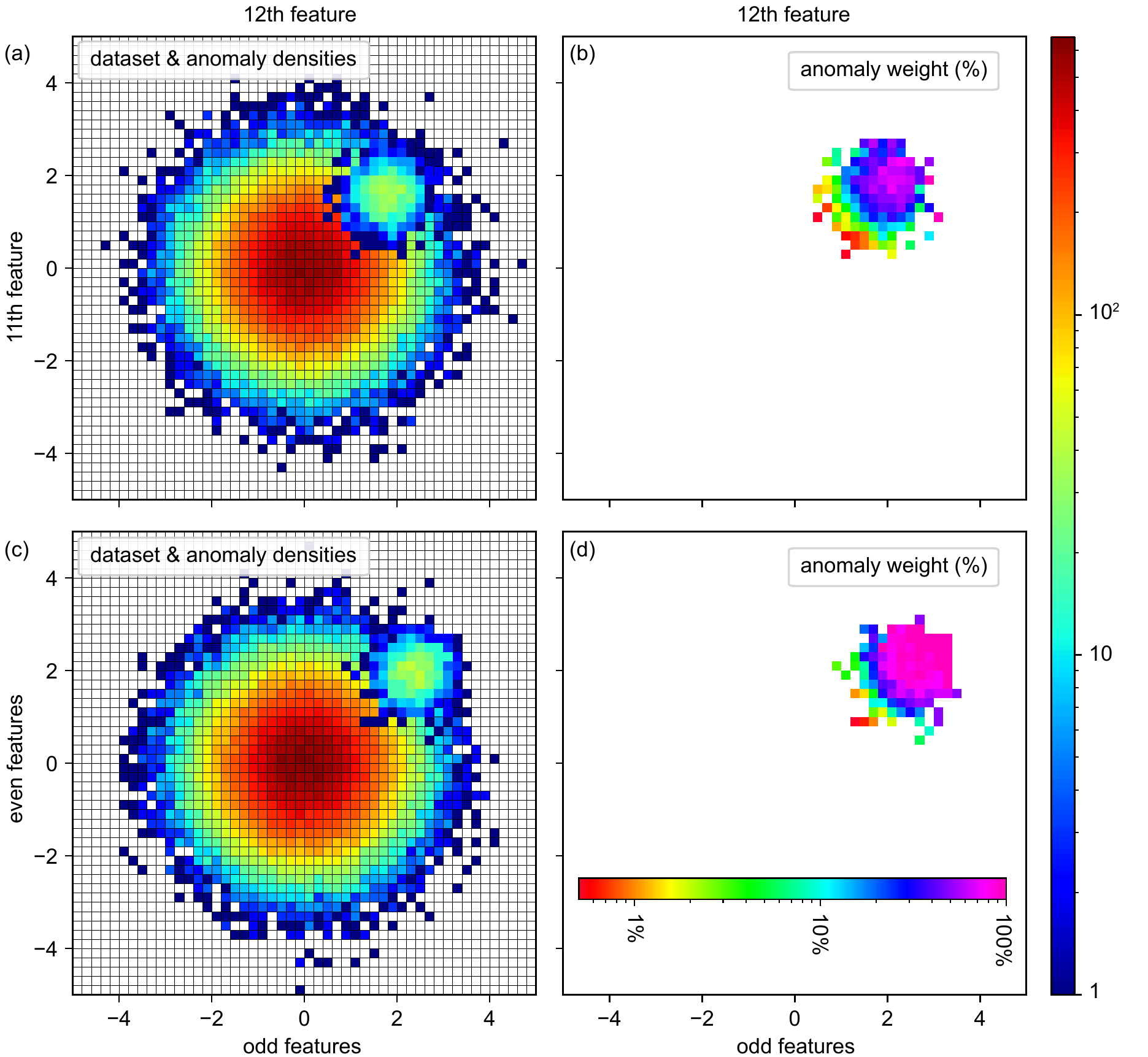}
      \caption{Dataset and anomaly histograms using only anomaly-revealing axes: the 11th feature versus the 12th (a) and even versus odd features (c). Corresponding anomaly weights in the overall distributions (b, d).}
      \label{fig:bells}
    \end{figure}

    It is necessary to combine several features in order to see anomalies stand up from the bell-shaped distribution of all values. In fig.~\ref{fig:bells}, the two axes with most eccentric anomaly distributions define a plane where anomalies are distinguishable from regular points. In the same manner, anomalies stand out when combining all the twelve anomaly-revealing axes, and their hyperplane is illustrated in fig.~\ref{fig:bells}c. This hyperplane is constructed by adding and normalizing to unitary standard deviation ($\sigma = 1$) even features and odd features respectively. Warning: Sum is conceptually incorrect when visualizing mse-based radii, and should be replaced in that case with the root-sum of squared features: $\sum_i x_i \rightarrow \sqrt{\sum_i x_i^2}$ when $x_C=0$. However, intuition can be built on this visualization, too.

    \textbf{Remark.} A simple and effective prediction can be made by looking at the positivity of feature values. Indeed, anomalies tend to present only positive values on axes from 1 to 12. The best AUROC score, $0.9864$, is obtained by \emph{simultaneous positiveness for features $7:12$}.

  \subsection{Dataset-derived baselines}
    Standalone radii-based classification (subsection \ref{sub:radial_class}) in the constructed dataset, $\mathcal{D}$, provides a series of baselines to be outperformed by the investigated methods. Given the symmetry of the present dataset, we choose to analyze radii from its center of mass, $x_C=0$. Having inside knowledge of the dataset---namely, the anomaly-revealing features are the first twelve and increase in anomaly distinguishability along their rank ($k=1..12$)---we consider the subspaces $\mathcal{D}_{k:12}$ given by all the consecutive axes between that include the 12th. All these twelve subspaces have the same center of mass, $x_C=0$. When we operate a radii-based classification in $\mathcal{D}_{i:12}$, we build a receiver operating characteristic (ROC curve), which plots the true-positive rate versus the false-positive rate when one varies the threshold-radius that separates anomalies from regular points. We designate these curves by the name \emph{radial baselines}. Fig.~\ref{fig:rocs} a displays ROCs for $\mathcal{D}_{1:12}$, $\mathcal{D}_{9:12}$, $\mathcal{D}_{12:12}$. The areas under these curves (AUROCs) measure the performance of the classification. Fig.~\ref{fig:rocs} b displays AUROCs for all twelve radial baselines yielded by $\mathcal{D}_{k:12}$ and shows that the best performing comes from $\mathcal{D}_{9:12}$. Therefore, including anomaly-less-revealing features (here, features up to the eighth) in calculations can be detrimental in anomaly detection. In the search of an even better baseline, we also investigate subspaces with less features ($\mathcal{D}_{9:11}$, $\mathcal{D}_{9:10}$, $\mathcal{D}_{9:9}$), but without observing a performance increase. Including axes on which anomalies follow the dataset distribution (features beyond the 12th) decrease AUROCs, as expected. In conclusion, the best radial baseline come from features $9:12$, with $\textrm{AUROC}=0.9392$.

  \subsection{Centric autoencoders in action}
\begin{figure}
	  \makebox[0.99 \textwidth]{ \includegraphics[width=0.7\textwidth]{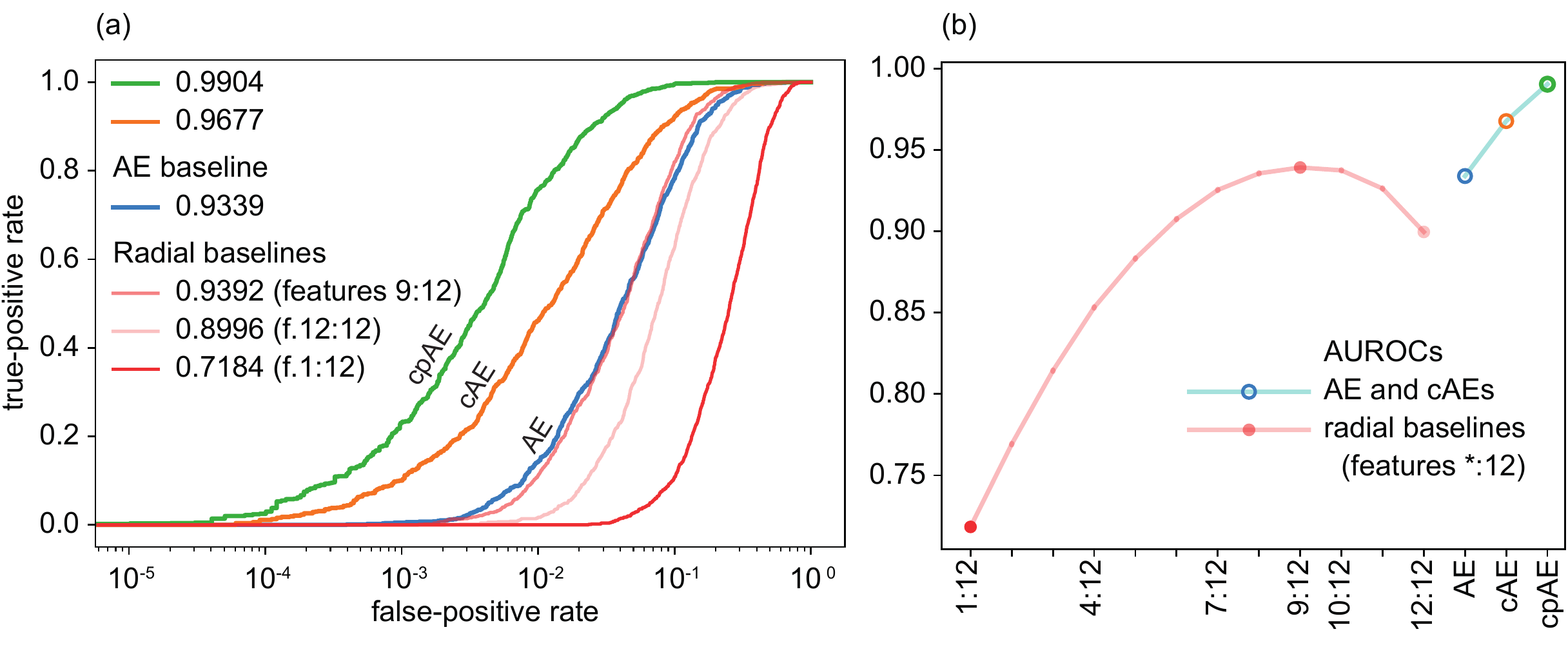} }
      \caption{(a) ROC curves for cpAE, cAE, AE (autoencoder baseline) and standalone radii-based classifications (radial baselines). Legend numbers correspond to areas under these curves. (b) Comparison between areas under ROCs, for AEs in (a) and for different selections of anomaly-revealing features.}
      \label{fig:rocs}
    \end{figure}

In this paper, autoencoders have 70 neurons in the encoder hidden layer, 8 neurons in the coding layer, and 70 neurons in the decoder hidden layer. Inputs and outputs have 15 units each. When a dataset center is needed, we assign it to the mass center of the training set, hence it is close to 0. First, we train an autoencoder (AE) and classify the dataset points by their reconstruction errors. A ROC curve is drawn for this classification (fig.~\ref{fig:rocs}) and it will serve as an extra baseline. Second, another autoencoder is trained and simultaneously tracked for the test set AUROC issued from the radii-based classification. The result of this process is a cAE (subsection \ref{sub:bare}). Its ROC curve is already better than the baselines (fig.~\ref{fig:rocs}). Next, building a cpAE (subsection \ref{sub:common}), in which the training set is compressed by a centric factor, $c$, is more complicated since one needs to efficiently explore the values of hyperparameter $c$. For this, we opt for a scalar centric factor and realize a Bayesian search via tree-structured Parzen estimators (\cite{bergstra}, \cite{bergstra11}, \cite{bergstra2013making}). Each step of the Bayesian search is equivalent to a full cAE training like above. The cpAE resulted in this process is a highly improved AE (fig.~\ref{fig:rocs} a, b). 

    There is temptation to boost further the cpAE by adopting a vector centric factor, in order to ensure differentiated compression on the 15 axes of the dataset. Besides the immensely higher cost of searching in the new hyperparameter space, that is a theoretically incorrect approach, as it combines two less expensive aspects: radial optimization and uniform compression.

    The way to outperform a cpAE is therefore post-training radial deformation of outputs and expansion of inputs. Hence, we apply the described radial optimization algorithm(s) both on the cAE and on the cpAE. 
    
\section{Results by dataset}
In the field, we have to adapt our instruments to practical aspects. We could see that sometimes feature selection improves the prediction ability of our centric autoencoders. The simplest way to select features is to see how they correlate with the truth labels, but also among themselves. This was applied to Kaggle1 (public credit card transaction dataset from Kaggle website). Another way is to classify along individual features and keep only the ones with the highest scores. 



\textbf{Results for all datasets}

In the following table, we group AUROC scores for all the datasets. The radial deformation is performed with two consecutive passes of our greedy algorithm and maximum 15 epochs of angular gradient ascent. Most vanilla autoencoder scores originate from other calculations, but are added for completeness. (The classification criterion of the vanilla AEs is the reconstruction error.)

\begin{center}
\begin{tabular}{ |p{3cm}||p{3.cm}|p{1cm}|p{1.cm}|p{1.cm}|p{1.cm}|p{1.cm} | }
  \hline
  \multicolumn{7}{|c|}{AUROC score in \%} \\
  \hline
  Method name & Artificial dataset & Kaggle1 & Retail1 & Retail2 & Retail3 & Retail4 \\
  \hline \hline
  Radial baseline		& 93.90		& 95.52	& 95.59	& 96.73  &  99.83		& 98.98		 \\
  Radial deform  		& 94.70		& 96.78	& 99.81	& 99.82  &  99.89		& 99.64		 \\
  \hline
  Vanilla AE    		& 93.39		& >95.4	& 95.30	& 97.11  &  95.42		& 94.60		  \\
  cAE           		& 96.77		& 96.42	& 98.72	& 97.21  &  99.84		& 99.06		 \\
  EIcAE         		& 99.03		& 96.46	& 98.75	& 98.68  &  99.84		& 99.08	 \\
  \hline
  DOcAE         		& 99.54		& 96.38	& 99.11	& 99.41  &  99.85		& 99.36	   \\
  \hline
 \end{tabular}
 \end{center}

Before calculations, Kaggle1 underwent the following feature selection: Amount, V1, V2, V3, V4, V5, V6, V7, V9, V10,
V11, V12, V14, V16, V17, V18, V19, V21.


The retained features are the columns: 50, 30, 38, 21, 9, 12, 52, 37, 36, 32, 41, 35. On each of these columns, distance from its average value gives a high AUROC score. In other terms, these are the most fraud-eccentric columns. In general, the classification scores for the artificial dataset are higher than for the others datasets. One reason is its particular structure. Another important reason is that we did a sustained hyperparameter search (with Hyperopt) for the artificial dataset cAEs, but not in the case of the other datasets. This was an assumed choice, in order to avoid doubts about the unsupervised character of the simple cAE and the EIcAE flavors. Scores for some \textbf{extra methods} on the artificial dataset are following. These methods are based on a cpAE instead of cAE. Their names are easy to understand.

\begin{center}
\begin{tabular}{ |p{2cm}||p{2cm}|  }
  \hline
  \multicolumn{2}{|c|}{ Artificial dataset } \\
  \hline
  Extra method  & AUROC  in \% \\
  \hline 
  \hline
  cpAE    		 & 99.04   \\
  EIcpAE  		 & 99.04   \\
  \hline
  DOcpAE  	 	 & 99.70   \\
  \hline
 \end{tabular}
 \end{center}
 
Comparing this cpAE score with the previous-table artificial-dataset EIcAE number, we found a situation where the two cAE types are equivalent.

\subsection{Radial deformations versus the basin-hopping algorithm}
\begin{figure}[!ht]
\makebox[0.99 \textwidth]{ \includegraphics[width=0.8\textwidth]{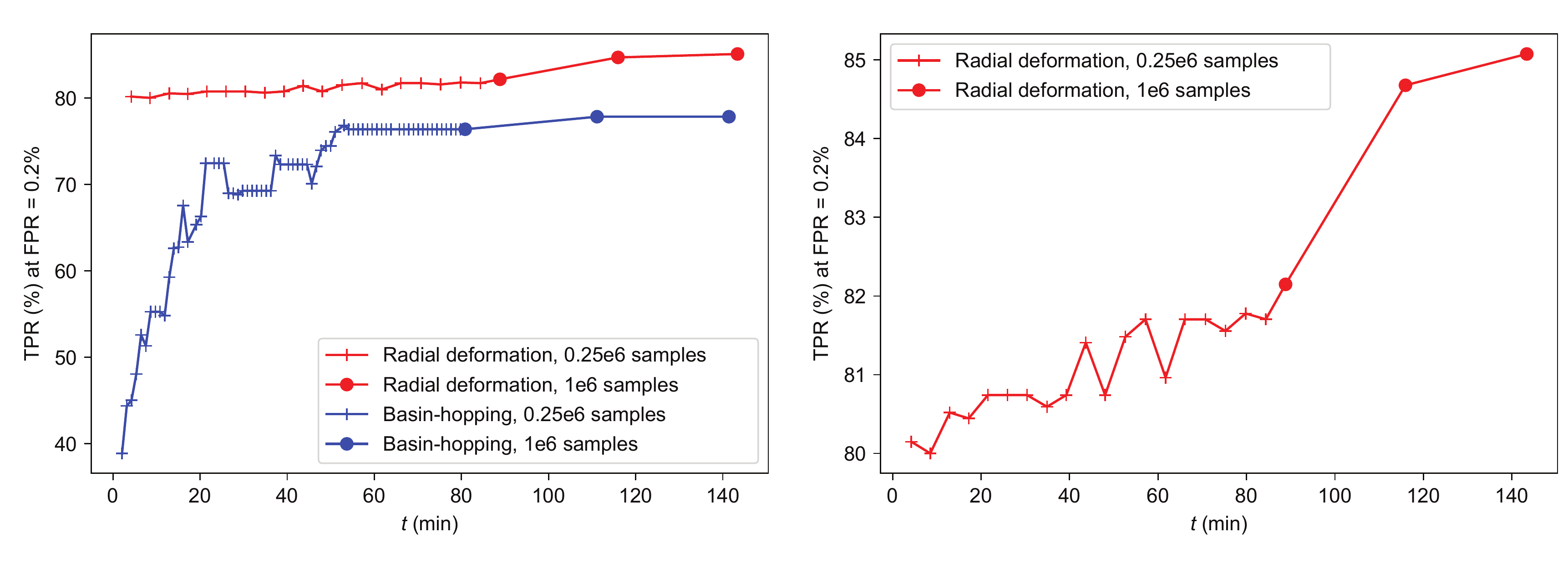} }
\caption{Radial deformation and basin-hopping, score versus time. One point represents 20 iterations of basin-hopping or two greedy search passes and maximum 15 epochs of angular gradient ascent. The underlying metric that basin-hopping tries to minimize is the sum of fraud ranks, in a sorted array where frauds are pushed toward lower ranks. The right plots zooms in the radial deformation curve.}\label{fig:basinhop}
\end{figure}

In this project, the development of the radial deformation algorithms brought us to compare them to the 
 basin-hopping method, that calculates feature weights in order to improve fraud detection. In fig.~\ref{fig:basinhop} we show face-to-face how the TPR(FPR=0.2\%) score evolves in time for the two approaches.

Radial deformation clearly outperforms basin-hopping---at least for this particular dataset, Retail1. Indeed, within several minutes, it is capable of reaching a higher score than the basin-hopping plateau that basin-hopping hits after one hour. This is a remarkable result for a method that was developed in order to complete the context of centric autoencoders.

\section{Conclusion}
  \label{conclusion}
This study proposes a novel approach to classification, built on radii-enriched autoencoders, that we term \emph{centric autoencoders}. The essence of the new objects that are proposed here is not anymore reconstruction, but a transformation of the dataset points, which conserves their dimensionality. Enjoying identical dimensionality between inputs and outputs allows the use of distances, more precisely radii from an established center, in the classification of inputs. The optimized version of centric autoencoders, the centripetal autoencoder, has an often-equivalent version that replaces the expensive search of compression factors at training time by the fast testing of expansion factors after training. Optimized centric autoencoders outperform convincingly classical autoencoders. They also offer consistent classification improvements over radial baselines, in the majority of the available datasets. Centric autoencoders are to be combined with radial deformation to further enhance anomaly detection. Last but not least, the specific advantage of keeping autoencoders in live detectors is their intrinsic ability to recognize previously unknown types of frauds. 

\footnote{We acknowledge NetGuardians' in providing access to the necessary banking data, as well as InnoSuisse's financial participation in the research project.}

\newpage
\bibliography{totocopy.bib}
\bibliographystyle{icml2019}


\end{document}